\documentclass[runningheads]{llncs}

\usepackage[T1]{fontenc}

\usepackage{authblk}
\usepackage{marvosym}
\usepackage{algorithm}
\usepackage{algpseudocode}
\usepackage{graphicx}
\usepackage{amsmath}
\usepackage{setspace}
\usepackage{bm}
\usepackage{multirow}
\usepackage{amsfonts}
\usepackage{color}
\usepackage{hyperref}
\hypersetup{
    colorlinks=true,
    linkcolor=blue,
    filecolor=magenta,      
    urlcolor=cyan,
    pdftitle={IFE},
    pdfpagemode=FullScreen,
    }

\graphicspath{ {pictures/} }

\newcommand{\ourmodel}{IFE}

\begin{document}

\title{Instructive Feature Enhancement for Dichotomous Medical Image Segmentation}

\titlerunning{IFE for Dichotomous Medical Image Segmentation}

\author{Lian Liu\inst{1,2,3}\thanks{Lian Liu and Han Zhou contributed equally to this work.}  \and
Han Zhou\inst{1,2,3}$^\star$  \and
Jiongquan Chen\inst{1,2,3} \and
Sijing Liu\inst{1,2,3} \and
Wenlong Shi\inst{4} \and
Dong Ni\inst{1,2,3} \and
Deng-Ping Fan\inst{5}\textsuperscript{(\Letter)} \and
Xin Yang\inst{1,2,3}\textsuperscript{(\Letter)} }

\authorrunning{L. Liu, H. Zhou et al.}

\institute{
National-Regional Key Technology Engineering Laboratory for Medical Ultrasound, School of Biomedical Engineering, Health Science Center, Shenzhen University, China\\
\email{xinyang@szu.edu.cn}\\
\and
Medical Ultrasound Image Computing (MUSIC) Lab, Shenzhen University, China
\and Marshall Laboratory of Biomedical Engineering, Shenzhen University, China
\and Shenzhen RayShape Medical Technology Co., Ltd, China
\and Computer Vision Lab (CVL), ETH Zurich, Zurich, Switzerland\\
\email{dengpfan@gmail.com}}

\maketitle              

\begin{abstract}
Deep neural networks have been widely applied in dichotomous medical image segmentation (DMIS) of many anatomical structures in several modalities, achieving promising performance. However, existing networks tend to struggle with task-specific, heavy and complex designs to improve accuracy. They made little instructions to which feature channels would be more beneficial for segmentation, and that may be why the performance and universality of these segmentation models are hindered. In this study, we propose an instructive feature enhancement approach, namely \textbf{\ourmodel}, to adaptively select feature channels with rich texture cues and strong discriminability to enhance raw features based on local curvature or global information entropy criteria. Being plug-and-play and applicable for diverse DMIS tasks, \ourmodel~encourages the model to focus on texture-rich features which are especially important for the ambiguous and challenging boundary identification, simultaneously achieving simplicity, universality, and certain interpretability. To evaluate the proposed IFE, we constructed the first large-scale DMIS dataset \textbf{Cosmos55k}, which contains 55,023 images from 7 modalities and 26 anatomical structures. Extensive experiments show that \ourmodel~can improve the performance of classic segmentation networks across different anatomies and modalities with only slight modifications. Code is available at \url{https://github.com/yezi-66/IFE}
\end{abstract}

\begin{figure}
\includegraphics[width=0.95\textwidth]{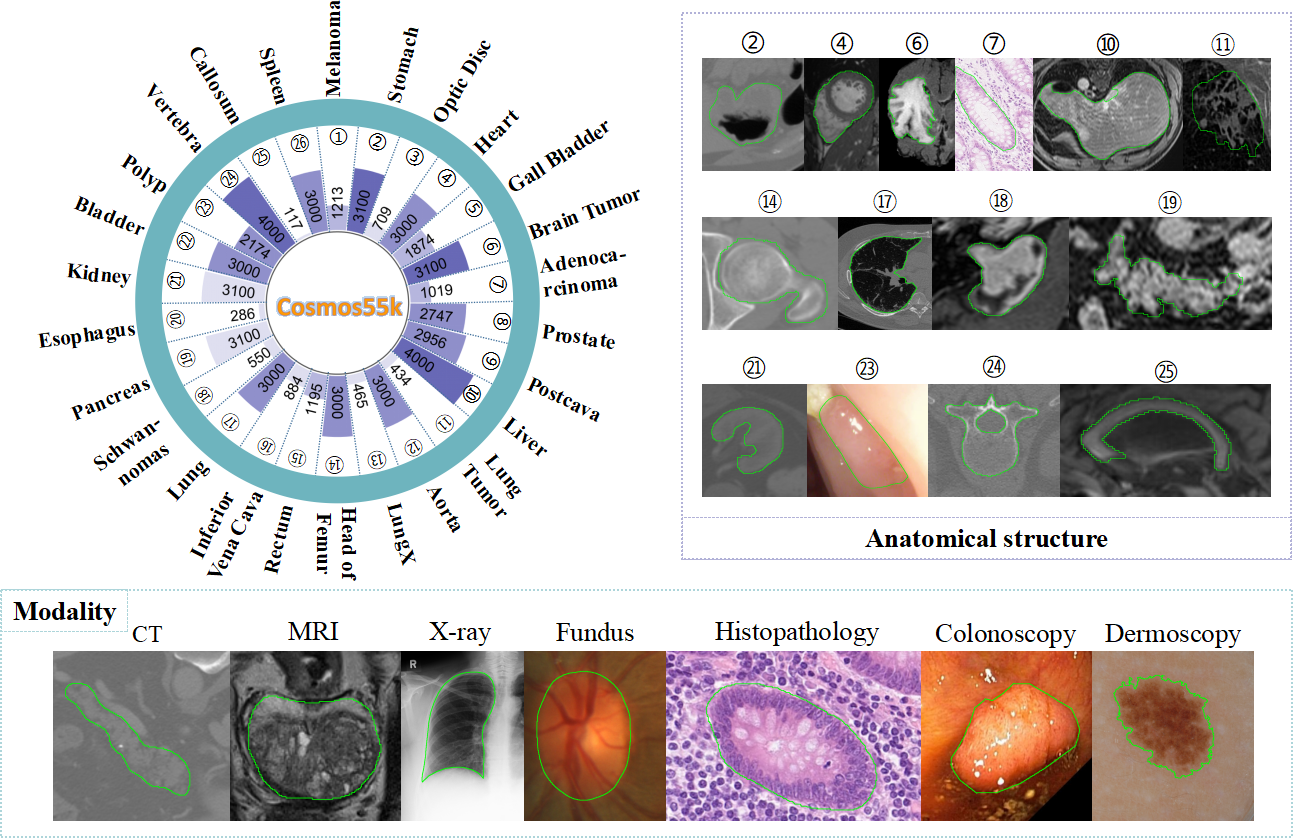}
\caption{Statistics, modalities, and examples of anatomical structures in Cosmos55k.} 
\label{fig1}
\end{figure}

\section{Introduction}
Medical image segmentation (MIS) can provide important information regarding anatomical or pathological structural changes and plays a critical role in computer-aided diagnosis. With the rapid development of intelligent medicine, MIS involves an increasing number of imaging modalities, raising requirements for the accuracy and universality of deep models. \par

Medical images have diverse modalities owing to different imaging methods. Images from the same modality but various sites can exhibit high similarity in overall structure but diversity in details and textures. Models trained on specific modalities or anatomical structure datasets may not adapt to new datasets. Similar to dichotomous image segmentation tasks~\cite{qin2022}, MIS tasks typically input an image and output a binary mask of the object, which primarily relies on the dataset. To facilitate such dichotomous medical image segmentation (DMIS) task, we constructed Cosmos55k, a large-scale dataset of 55,023 challenging medical images covering 26 anatomical structures and 7 modalities (see Fig.~\ref{fig1}). \par

Most current MIS architectures are carefully designed. Some increased the depth or width of the backbone network, such as UNet++~\cite{zhou2019unet++}, which uses nested and dense skip connections, or DeepLabV3+~\cite{chen2018encoder}, which combines dilated convolutions and feature pyramid pooling with an effective decoder module. Others have created functional modules such as Inception and its variants~\cite{szegedy2017inception,chollet2017xception}, depthwise separable convolution~\cite{howard2019searching}, attention mechanism~\cite{hou2021coordinate,zhong2020squeeze}, and multi-scale feature fusion~\cite{chen2017deeplab}. Although these modules are promising and can be used flexibly, they typically require repeated and handcrafted adaptation for diverse DMIS tasks. Alternatively, frameworks like nnUNet~\cite{isensee2021nnu} developed an adaptive segmentation pipeline for multiple DMIS tasks by integrating key dataset attributes and achieved state-of-the-art. However, heavy design efforts are needed for nnUNet and the cost is very expensive. More importantly, previous DMIS networks often ignored the importance of identifying and enhancing the determining and instructive feature channels for segmentation, which potentially limits their performance in general DMIS tasks. \par

We observed that the texture-rich and sharp-edge cues in specific feature channels are crucial and instructive for accurately segmenting objects. Curvature~\cite{gong2017curvature} can represent the edge characteristics in images. Information entropy~\cite{abdel2017two} can describe the texture and content complexity of images. In this study, we focus on exploring their roles in quantifying feature significance. Based on curvature and information entropy, we propose a simple approach to balance accuracy and universality with only minor modifications of the classic networks. Our contribution is three folds. First, we propose the novel 2D DMIS task and construct a large-scale dataset (\textit{Cosmos55k}) to benchmark the goal and we will release the dataset to contribute to the community. Second, we propose a simple, generalizable, and effective instructive feature enhancement approach (\textit{\ourmodel}). With extensive experiments on Cosmos55k, IFE soundly proves its advantages in promoting various segmentation networks and tasks. Finally, we provide an interpretation of which feature channels are more beneficial to the final segmentation results. We believe IFE will benefit various DMIS tasks. \par

\begin{figure}
\centering
\includegraphics[width=0.98\textwidth]{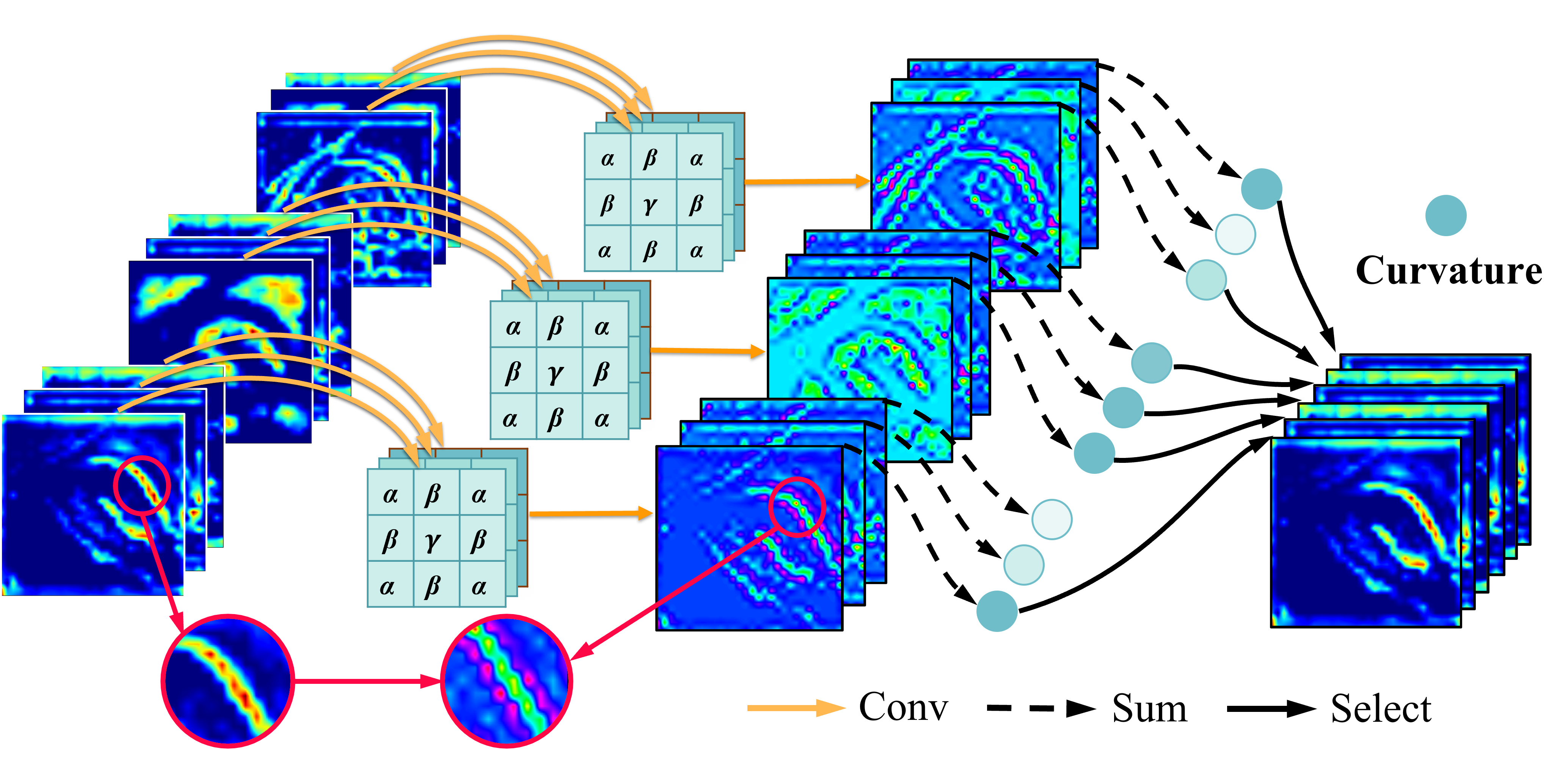}
\caption{Example of feature selection using curvature.} 
\label{cur}
\end{figure}

\section{Methodology}
\begin{figure}
\centering
\includegraphics[width=0.98\textwidth]{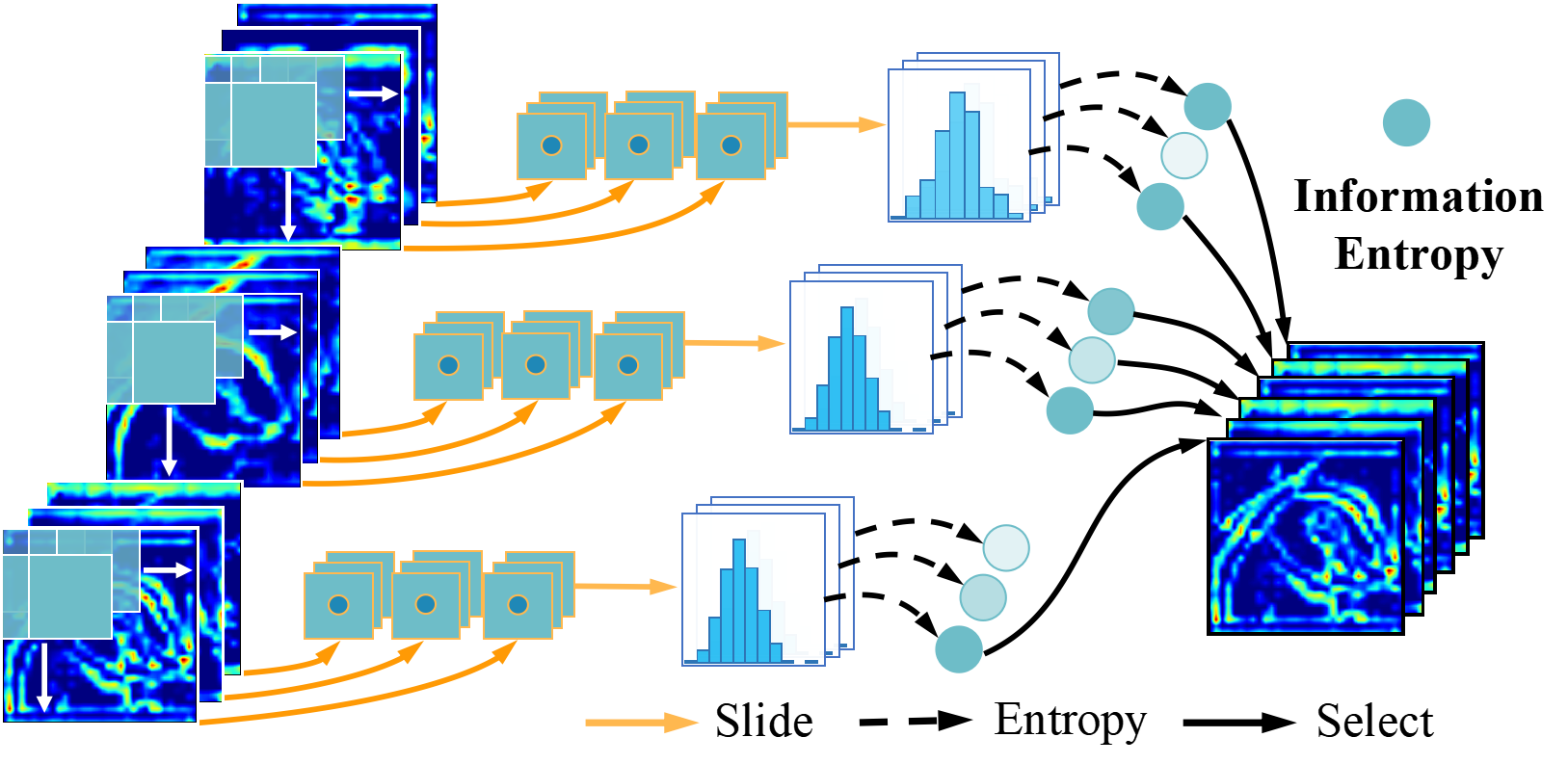}
\caption{Example of feature selection using 2D information entropy.} 
\label{entro}
\end{figure}

\textbf{Overview.}
Fig.~\ref{cur} and Fig.~\ref{entro} illustrate two choices of the proposed \ourmodel. It is general for different DMIS tasks. \textbf{1)} We introduce a feature quantization method based on either curvature (Fig.~\ref{cur}) or information entropy (Fig.~\ref{entro}), which characterizes the content abundance of each feature channel. The larger these parameters, the richer the texture and detail of the corresponding channel feature. \textbf{2)} We select a certain proportion of channel features with high curvature or information entropy and combine them with raw features. \ourmodel~improves performance with minor modifications to the segmentation network architecture. \par

\subsection{Curvature-based Feature Selection}
For a two-dimensional surface embedded in Euclidean space $R^{3}$, two curvatures exist: Gaussian curvature and mean curvature. Compared with the Gaussian curvature, the mean curvature can better reflect the unevenness of the surface. Gong~\cite{gong2017curvature} proposed a calculation formula that only requires a simple linear convolution to obtain an approximate mean curvature, as shown below:

\begin{equation}
\setlength{\arraycolsep}{3pt}
\bm{C} = \begin{bmatrix}
 \bm{C_{1}} & \bm{C_{2}} & \bm{C_{3}} \\[2pt]
\end{bmatrix} \bm{*} \bm{X},
\label{curvature}
\end{equation}

where $\bm{C_{1}}=[\alpha, \beta, \alpha]^{T}$, $\bm{C_{2}}=[\beta, \gamma, \beta]^{T}$, $\bm{C_{3}}=[\alpha, \beta, \alpha]^{T}$, the values of $\alpha$, $\beta$, and $\gamma$ are $-1/16$, $5/16$, $-1$. $\bm{*}$ denotes convolution, $\bm{X}$ represents the input image, and $\bm{C}$ is the mean curvature. 
Fig.~\ref{cur} illustrates the process, showing that the curvature image can effectively highlight the edge details in the features.\par

\subsection{Information Entropy-based Feature Selection}
As a statistical form of feature, information entropy~\cite{abdel2017two} reflects the spatial and aggregation characteristics of the intensity distribution. It can be formulated as:

\begin{equation}
\ E=-\sum^{255}_{i=0}\sum^{255}_{j=0}P_{i,j}log_{2}\left(P_{i,j}\right)\\[3pt], 
\ P_{i,j}=f\left(i_n,j_n\right)/\left(H\times{W}\right),
\label{entropy}
\end{equation}

$i_n$ denotes the gray value of the center pixel in the $n^{th}$ $3\times{3}$ sliding window and $j_n$ denotes the average gray value of the remaining pixels in that window (teal blue box in Fig.~\ref{entro}). The probability of $\left(i_n,j_n\right)$ occurring in the entire image is denoted by $P_{i,j}$, and $E$ represents the information entropy.\par

\begin{algorithm}[b]
\scriptsize
	\caption{Information entropy of features with histogram.} 
	\label{alg1} 
    \begin{spacing}{1.3}
	\begin{algorithmic}
		\State Input:$\boldsymbol{F_x} \in \mathbb{ R^{C\times{H}\times{W}}}, bins = 256, kernel\_size = 3$ 
        \State Output:$\bm{E} \in \mathbb{R^{C}}$
		\State  {$z = unfold\left(\boldsymbol{F}, kernel\_size\right)$}  \textcolor{blue}{\Comment{Sliding window operation.}}
		\State  $i = flatten\left(z\right)\left[\left({H\times{W}}\right)//2\right]$
		\State  $j = \left(sum\left(z\right) - i\right)/{\left(\left({H\times{W}}\right)-1\right)}$
		\State  $f_{hist\left(i,j\right)} = histc\left(\left(i, j\right),bins\right) $ \textcolor{blue}{\Comment{Compute the histogram of $\left(i, j\right)$.}}
        \State  $ext\_k= kernel\_size // 2$ 
        \State $\bm{P_{hist\left(i,j\right)}}=f_{hist\left(i,j\right)} / \left(\left(H+ext_k\right)\times{\left(W+ext_k\right)}\right)$
		\State  $\bm{E}=sum\left(-\bm{P_{hist\left(i,j\right)}}\times{log_2\left(\bm{P_{hist\left(i,j\right)}}\right)}\right)$
	\end{algorithmic} 
    \end{spacing}
\end{algorithm}

\begin{figure}[t!]
\centering
\includegraphics[width=.98\textwidth]{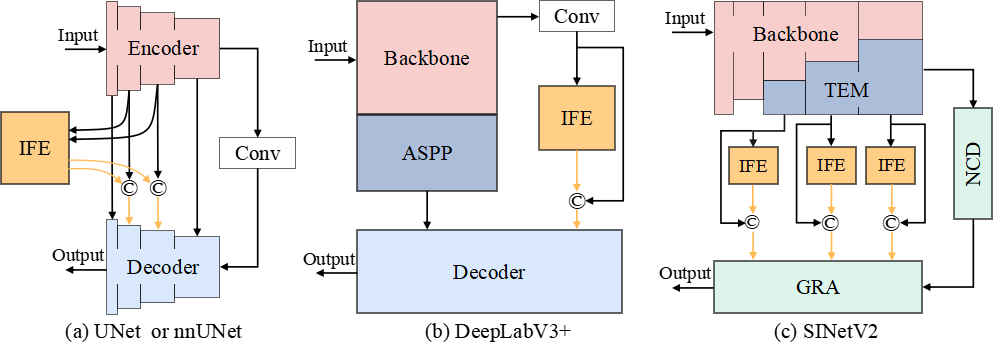}
\caption{Implementation of \ourmodel~in exemplar networks. $\copyright$ is concatenation.}
\label{fig3}
\end{figure}

Each pixel on images corresponds to a gray or color value ranging from 0 to 255. However, each element in the feature map represents the activation level of the convolution filter at a particular position in the input image. Given an input feature $\bm{F_x}$, as shown in Fig.~\ref{entro}, the tuples $\left(i,j\right)$ obtained by the sliding windows are transformed to a histogram, representing the magnitude of the activation level and distribution within the neighborhood. This involves rearranging the activation levels of the feature map. The histogram converting method $histc$ and information entropy $\bm{E}$ are presented in Algorithm.~\ref{alg1}. Note that the probability 
$\bm{P_{hist\left(i,j\right)}}$ will be used to calculate the information entropy.\par

\subsection{Instructive Feature Enhancement}
Although \ourmodel~can be applied to various deep neural networks, this study mainly focuses on widely used segmentation networks. Fig.~\ref{fig3} shows the framework of \ourmodel~embedded in representative networks, \textit{e.g.}, DeepLabV3+~\cite{chen2018encoder}, UNet~\cite{ronneberger2015u}, nnUNet~\cite{isensee2021nnu}, and SINetV2~\cite{fan2021concealed}. 
The first three are classic segmentation networks. Because the MIS task is similar to the camouflage object detection, such as low contrast and blurred edges, we also consider SINetV2~\cite{fan2021concealed}. 
According to \cite{raghu2021vision}, we implement \ourmodel~on the middle layers of UNet~\cite{ronneberger2015u} and nnUNet~\cite{isensee2021nnu}, on the low-level features of DeepLabV3+~\cite{chen2018encoder}, and on the output of the TEM of SINetV2~\cite{fan2021concealed}.\par

\begin{figure}[t!]
\centering
\includegraphics[width=.98\textwidth]{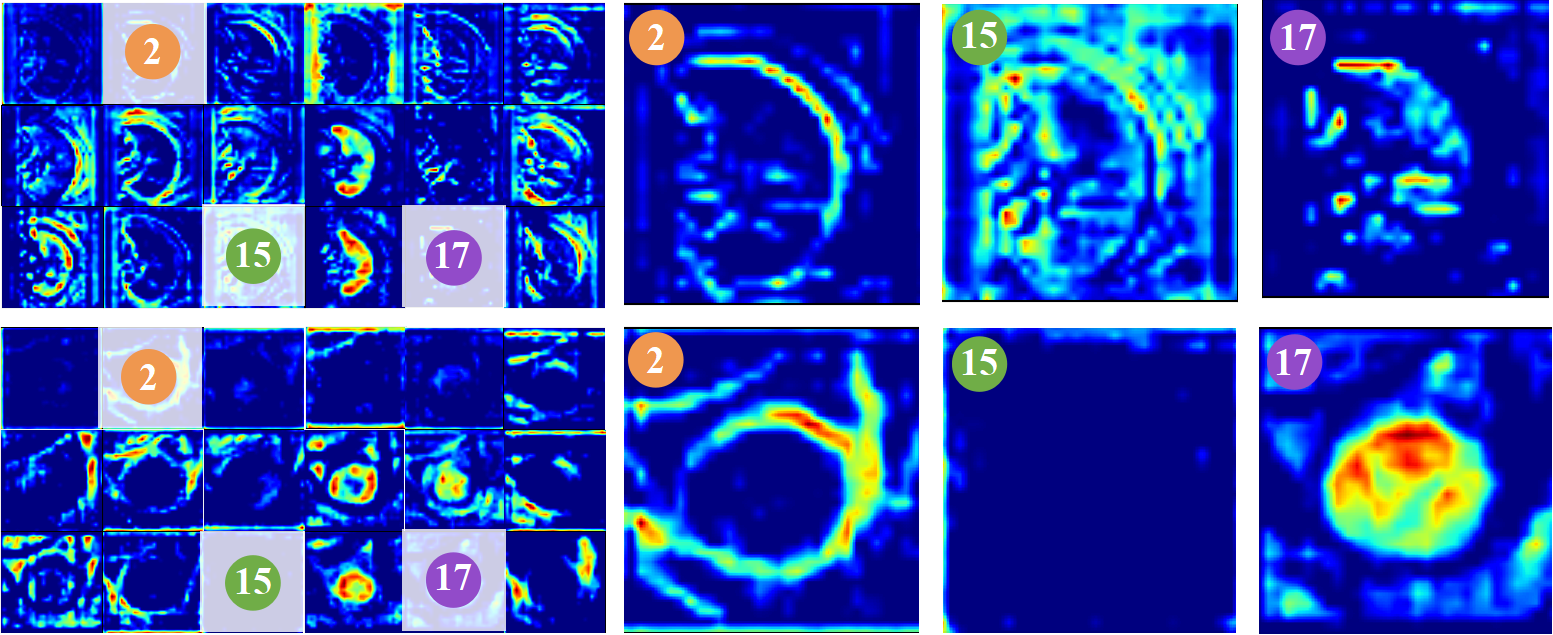}
\caption{Feature maps visualization of the lung (first row) and aortic (second row) CT images from stage 3 of SINetV2. 2, 15, and 17 are the indexes of channels. The information contained by the same channel may vary across different images.} 
\label{fig4}
\end{figure}

While the input images are encoded into the feature space, the different channel features retain textures in various directions and frequencies. Notably, the information contained by the same channel may vary across different images, which can be seen in Fig.~\ref{fig4}. For instance, the $15^{th}$ channel of the lung CT feature map contains valuable texture and details, while the same channel in the aortic CT feature map may not provide significant informative content. However, their $2^{nd}$ channel features both focus on the edge details. By preserving the raw features, the channel features that contribute more to the segmentation of the current object can be enhanced by dynamically selecting from the input features. Naturally, it is possible to explicitly increase the sensitivity of the model to the channel information. Specifically, 
for input image $\mathbf{X}$, the deep feature $ \boldsymbol{F_x} = [f_1, f_2, f_3,\cdots,f_C]\in \mathbb{R^{C*H*W}}$ can be obtained by an encoder with the weights $\theta_x$: $\boldsymbol{F_x}=Encoder\left(\mathbf{X},\theta_x\right) $, our \ourmodel~can be expressed as:

\begin{equation}
\boldsymbol{F_{x}^{'}} = max\{S\left(\boldsymbol{F_x}\right),r\},
\end{equation}

$\boldsymbol{F_{x}^{'}}$ is the selected feature map and $S$ is the quantification method (see Fig.~\ref{cur} or Fig.~\ref{entro}), and $r$ is the selected proportion. As discussed in~\cite{cao2018feature}, enhancing the raw features through pixel-wise addition may introduce unwanted background noise and cause interference. In contrast, the concatenate operation directly joins the features, allowing the network to learn how to fuse them automatically, reducing the interference caused by useless background noises. Therefore, we used the concatenation and employed the concatenated features $\boldsymbol{F} = [\boldsymbol{F_x},  \boldsymbol{F_{x}^{'}}]$ as the input to the next stage of the network. Only the initialization channel number of the corresponding network layer needs to be modified. \par

\section{Experimental Results}
\subsubsection{Cosmos55k.} 
To construct the large-scale Cosmos55k, 30 publicly available datasets~\cite{simpson2019large,zhou2019prior,litjens2014evaluation,zhao2022coarse,ji2021learning,kavur2021chaos,bernard2018deep,bakas2018identifying,jha2020kvasir,codella2018skin,ji2022amos,luo2022word,sekuboyina2021verse} were collected and processed with organizers' permission. The processing procedures included uniform conversion to PNG format, cropping, and removing mislabeled images. Cosmos55k (Fig.~\ref{fig1}) offers 7 imaging modalities, including CT, MRI, X-ray, fundus, etc., covering 26 anatomical structures such as the liver, polyp, melanoma, and vertebra, among others. Images contain just one labeled object, reducing confusion from multiple objects with different structures. \par

\begin{table}
\scriptsize
\setlength{\tabcolsep}{3.2pt}
\caption{Quantitative comparison on \ourmodel. +C, +E means curvature- or information entropy-based \ourmodel. DLV3+ is DeepLabV3+. \textbf{Bolded} means the best group result. $*$ denotes that DSC passed \textit{t-test, p < 0.05}.}
\label{results}
\begin{tabular}{l c c c c c c c c c c}
\hline
Method & \textit{Con(\%)}$\uparrow$ & \textit{DSC(\%)}$\uparrow$ & \textit{JC(\% )}$\uparrow$ & \textit{F1(\%)}$\uparrow$ & \textit{HCE}$\downarrow$ & \textit{MAE(\%)}$\downarrow$ & \textit{HD}$\downarrow$ & \textit{ASD}$\downarrow$ & \textit{RVD}$\downarrow$\\
\hline
UNet & 84.410 & 93.695 & 88.690 & \textbf{94.534} & 1.988 & \textbf{1.338} & 11.464 & 2.287 & 7.145\\
+C & 85.666& $94.003^*$ & 89.154 & 94.528 & 1.777 & 1.449 & \textbf{11.213} & 2.222 & 6.658\\
+E & \textbf{86.664} & $\textbf{94.233}^*$ & \textbf{89.526} & 94.466 & \textbf{1.610} & 1.587 & 11.229 & \textbf{2.177} & \textbf{6.394}\\
\hline
nnUNet & -53.979 & 92.617 & 87.267 & 92.336 & 2.548 & 0.257 & 19.963 & 3.698 & 9.003\\
+C & \textbf{-30.908} & \textbf{92.686} & \textbf{87.361} & \textbf{92.399} & 2.521 & 0.257 & 19.840 & \textbf{3.615} & 8.919\\
+E & -34.750 & 92.641 & 87.313 & 92.367 & \textbf{2.510} & 0.257 & \textbf{19.770} & 3.637 & \textbf{8.772}\\
\hline
SINetV2 & 80.824 & 93.292 & 88.072 & 93.768 & 2.065 & 1.680 & 11.570 & 2.495 & 7.612\\
+C & \textbf{84.883} & $\textbf{93.635}^*$ & \textbf{88.525} & \textbf{94.152} & \textbf{1.971} & \textbf{1.573} & \textbf{11.122} & \textbf{2.402} & \textbf{7.125}\\
+E & 83.655 & 93.423 & 88.205 & 93.978 & 2.058 & 1.599 & 11.418 & 2.494 & 7.492\\
\hline
DLV3+ & 88.566 & 94.899 & 90.571 & 95.219 & 1.289 & \textbf{1.339} & 9.113 & 2.009 & 5.603\\
+C & \textbf{88.943} & $95.000^*$ & 90.738 & \textbf{95.239} & 1.274 & 1.369 & \textbf{8.885} & \textbf{1.978} & \textbf{5.391}\\
+E & 88.886 & $\textbf{95.002}^*$ & \textbf{90.741} & 95.103 & \textbf{1.257} & 1.448 & 9.108 & 2.011 & 5.468\\
\hline
\end{tabular}
\end{table}

\subsubsection{Implementation Details.} 
Cosmos55k comprises 55,023 images, with 31,548 images used for training, 5,884 for validation, and 17,591 for testing. We conducted experiments using Pytorch for UNet~\cite{ronneberger2015u}, DeeplabV3+~\cite{chen2018encoder}, SINetV2~\cite{fan2021concealed}, and nnUNet~\cite{isensee2021nnu}. The experiments were conducted for 100 epochs on an RTX 3090 GPU. The batch sizes for the first three networks were 32, 64, and 64, respectively, and the optimizer used was Adam with an initial learning rate of $10^{-4}$. Every 50 epochs, the learning rate decayed to 1/10 of the former. Considering the large scale span of the images in Cosmos55k, the images were randomly resized to one of seven sizes (224, 256, 288, 320, 352, or 384) before being fed into the network for training. During testing, the images were resized to a fixed size of 224. Notably, the model set the hyperparameters for nnUNet~\cite{isensee2021nnu} automatically. \par

\begin{table}
\scriptsize
\setlength{\tabcolsep}{1.7pt}
\caption{Ablation studies of UNet about ratio $r$. \textbf{Bolded} means the best result.}
\label{Ablation studies}
\begin{tabular}{l c c c c c c c c c c}
\hline
Method & Ratio & \textit{Con(\%)}$\uparrow$ & \textit{DSC(\%)}$\uparrow$ & \textit{JC(\% )}$\uparrow$ & \textit{F1(\%)}$\uparrow$ & \textit{HCE}$\downarrow$ & \textit{MAE(\%)}$\downarrow$ & \textit{HD}$\downarrow$ & \textit{ASD}$\downarrow$ & \textit{RVD}$\downarrow$\\
\hline
\multirow{2}*{UNet}  &  0,0 & 84.410 & 93.695 & 88.690 & \textbf{94.534} & 1.988 & \textbf{1.338} & 11.464 & 2.287 & 7.145\\
& 1.0,1.0 & 84.950 & 93.787 & 88.840 & 94.527 & 1.921 & 1.376 & 11.417 & 2.271 & 6.974\\
\hline
\multirow{3}*{+C} & 0.75,0.75 & 84.979 & 93.855 & 88.952 & 94.528 & 1.865 & 1.405 & 11.378 & 2.246 & 6.875\\
& 0.75,0.50 & 85.666& 94.003 & 89.154 & 94.528 & 1.777 & 1.449 & \textbf{11.213} & 2.222 & 6.658\\
& 0.50,0.50 & 84.393 & 93.742 & 88.767 & 94.303 & 1.908 & 1.497 & 11.671 & 2.305 & 7.115\\
\hline
\multirow{3}*{+E} & 0.75,0.75 &85.392 & 93.964 & 89.106 & 94.290 & 1.789 & 1.597 & 11.461 & 2.252 & 6.712\\
& 0.75,0.50 & 83.803 & 93.859 & 88.949 & 94.420 & 1.877 & 1.471 & 11.351 & 2.260 & 6.815\\
& 0.50,0.50 &\textbf{86.664} & \textbf{94.233} & \textbf{89.526} & 94.466 & \textbf{1.610} & 1.587 & 11.229 & \textbf{2.177} & \textbf{6.394}\\
\hline
\end{tabular}
\end{table} 

\begin{figure}[t!]
\centering
\includegraphics[width=0.98\textwidth]{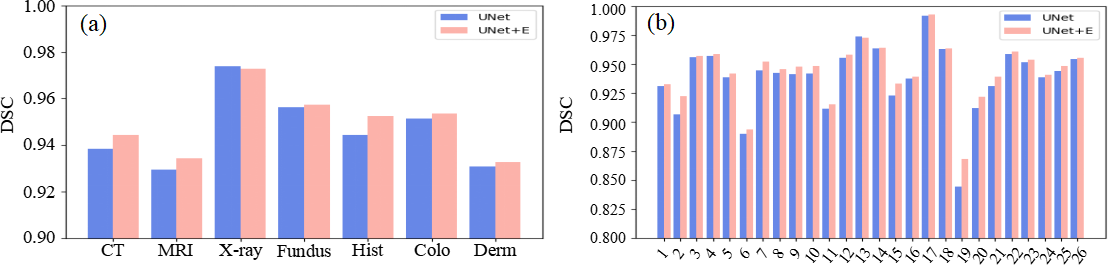}
\setlength{\abovecaptionskip}{-0.cm} 
\caption{DSC of UNet (blue) and UNet+E (pink) in modalities (a) and anatomic structures (b). Hist, Colo, and Derm are Histopathology, Colonoscopy, and Dermoscopy.} 
\label{hist_result}
\end{figure}

\begin{figure}[t!]
\centering
\includegraphics[width=0.98\textwidth]{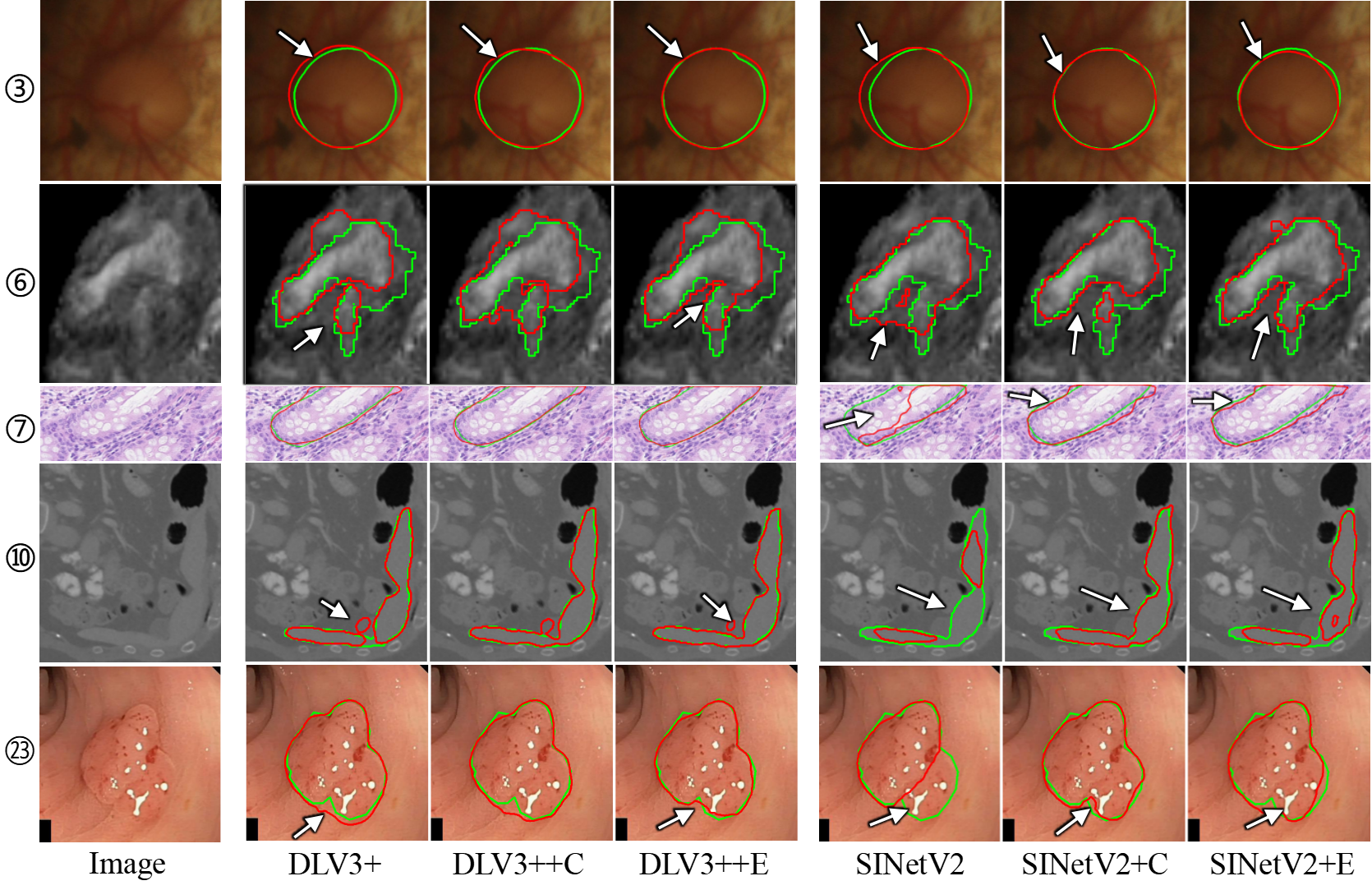}
\setlength{\abovecaptionskip}{-0.cm} 
\caption{Qualitative comparison of different models equipped with \ourmodel. Red and green denote prediction and ground truth, respectively.} 
\label{result}
\end{figure}

\subsubsection{Quantitative and Qualitative Analysis.} 
To demonstrate the efficacy of \ourmodel, we employ the following metrics: \textit{Conformity} (\textit{Con})~\cite{chang2009performance}, \textit{Dice Similarity Coefficient (DSC)}~\cite{bilic2023liver}, \textit{Jaccard Distance (JC)}~\cite{crum2006generalized}, \textit{F1}~\cite{achanta2009frequency}, \textit{Human Correction Efforts (HCE)}~\cite{qin2022}, \textit{Mean Absolute Error (MAE)}~\cite{6247743}, \textit{Hausdorff Distance (HD)}~\cite{zhou2019iou}, \textit{Average Symmetric Surface Distance (ASD)}~\cite{dubuisson1994modified}, \textit{Relative Volume Difference(RVD)}~\cite{dubuisson1994modified}. 
 The quantitative results for UNet~\cite{ronneberger2015u}, DeeplabV3+~\cite{chen2018encoder}, SINetV2~\cite{fan2021concealed}, and nnUNet~\cite{isensee2021nnu} are presented in Table~\ref{results}. From the table, it can be concluded that \ourmodel~can improve the performance of networks on most segmentation metrics. Besides,  Fig.~\ref{hist_result} shows that IFE helps models perform better in most modalities and anatomical 
 structures. Fig.~\ref{result} presents a qualitative comparison. \ourmodel~aids in locating structures in an object that may be difficult to notice and enhances sensitivity to edge gray variations. \ourmodel~can substantially improve the segmentation accuracy of the base model in challenging scenes.\par

\subsubsection{Ablation Studies.} 
Choosing a suitable selection ratio $r$ is crucial when applying \ourmodel~to different networks. Different networks' encoders are not equally capable of extracting features, and the ratio of channel features more favorable to the segmentation result varies. To analyze the effect of $r$, we conducted experiments using UNet~\cite{ronneberger2015u}. As shown in Table~\ref{Ablation studies}, either too large or too small $r$ will lead to a decline in the model performance.\par

\section{Conclusion}
In order to benchmark the general DMIS, we build a large-scale dataset called Cosmos55k. To balance universality and accuracy, we proposed an approach (\ourmodel) that can select instructive feature channels to further improve the segmentation over strong baselines against challenging tasks. Experiments showed that \ourmodel~can improve the performance of classic models with slight modifications in the network. It is simple, universal, and effective. Future research will focus on extending this approach to 3D tasks.\par

\subsubsection{Acknowledgements.}
The authors of this paper sincerely appreciate all the challenge organizers and owners for providing the public MIS datasets including 
AbdomenCT-1K, ACDC, AMOS 2022, BraTS20, CHAOS, CRAG, crossMoDA, EndoTect 2020, ETIS-Larib Polyp DB, iChallenge-AMD, iChallenge-PALM, IDRiD 2018, ISIC 2018, I2CVB, KiPA22, KiTS19$\&$KiTS21, Kvasir-SEG, LUNA16, Multi-Atlas Labeling Beyond the Cranial Vault (Abdomen), Montgomery County CXR Set, M\&Ms, MSD, NCI-ISBI 2013, PROMISE12, QUBIQ 2021, SIIM-ACR, SLIVER07, VerSe19$\&$VerSe20, Warwick-QU, and WORD. \par
This work was supported by the grant from National Natural Science Foundation of China (Nos. 62171290, 62101343), Shenzhen-Hong Kong Joint Research Program (No. SGDX20201103095613036), and Shenzhen Science and Technology Innovations Committee (No. 20200812143441001). 

\bibliographystyle{splncs04}
\bibliography{ref.bib}
\end{document}